\pdfoutput=1

\documentclass[11pt]{article}

\usepackage{EMNLP2022}

\usepackage{times}
\usepackage{latexsym}

\usepackage[T1]{fontenc}

\usepackage[utf8]{inputenc}

\usepackage{amsmath}
\usepackage{adjustbox}
\usepackage{graphicx}
\usepackage{makecell}

\usepackage{microtype}

\usepackage{inconsolata}

\usepackage[switch]{lineno}

\usepackage{subfiles}

\title{DynaMaR: Dynamic Prompt with Mask Token Representation}

\author{
	Xiaodi Sun$^1$, Sunny Rajagopalan$^{2,}$\thanks{$^\ast$Work done while at Amazon.} , Priyanka Nigam$^1$, Weiyi Lu$^1$, Yi Xu$^1$ \\
	{\bf Belinda Zeng}$^1$, {\bf Trishul Chilimbi}$^1$ \\
	$^1$Amazon, $^2$Google \\ $^1${\small \texttt{\{xiaodisu,nigamp,jazsingh,weiyilu,yxaamzn,zengb,trishulc\}@amazon.com}} \\ $^2${\small \texttt{sunny.rg@gmail.com}} \\ 
}

\begin{document}
\maketitle
\begin{abstract}
Recent research has shown that large language models pretrained using unsupervised approaches can achieve significant performance improvement on many downstream tasks. Typically when adapting these language models to downstream tasks, like a classification or regression task, we employ a fine-tuning paradigm in which the sentence representation from the language model is input to  a task-specific head; the model is then fine-tuned end-to-end. However, with the emergence of models like GPT-3, prompt-based fine-tuning has been proven to be a successful approach for few-shot tasks. Inspired by this work, we study discrete prompt technologies in practice. There are two issues that arise with the standard prompt approach. First, it can overfit on the prompt template. Second, it requires manual effort to formulate the downstream task as a language model problem. In this paper, we propose an improvement to prompt-based fine-tuning that addresses these two issues. We refer to our approach as DynaMaR -- {\bf Dyna}mic Prompt with {\bf Ma}sk Token {\bf R}epresentation. Results show that DynaMaR can achieve an average improvement of 10\% in few-shot settings and improvement of 3.7\% in data-rich settings over the standard fine-tuning approach on four e-commerce applications.
\end{abstract}

\section{Introduction}\label{sec:introduction}

Unsupervised pre-trained Language Models (LMs) such as BERT \citep{Devlin2019BERT} and RoBERTa \citep{Liu2019RoBERTaAR} have achieved state-of-the-art performance on many natural language understanding tasks. In general, these models are fine-tuned for different tasks through the addition of a task-specific head on top of the [CLS] token representation \citep{Scao2021HowMD}.

An alternative method to applying LMs on downstream tasks is through discrete prompts. A discrete prompt is an additional text phrase inserted along with the original input text that encapsulates the task of interest. By adding the prompt, we convert the downstream task into a masked language (MLM) problem. For example, to classify the sentiment of a movie review, ``I hate this movie.'', we can append a prompt to the input to get ``I hate this movie. It was [MASK]''. The pre-trained language model is thus prompted to identify the sentiment of the input statement and classify the [MASK] token as ``terrible'' instead of ``great'' \citep{Liu2021PretrainPA}. In this paper, we call a function that includes a prompt and its position information a prompt template.

\begin{figure}[t]
	\centering
	\includegraphics[width=.85\linewidth]{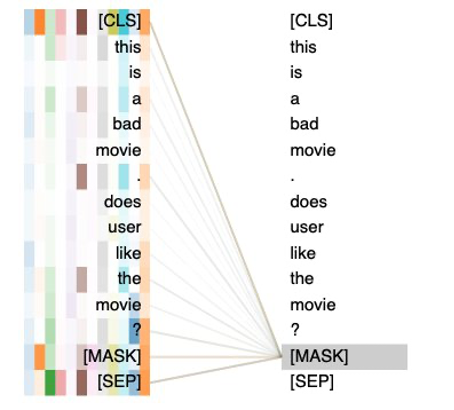}
	\caption{BERT Attention Distribution. The figure shows that the MLM model puts greater attention on the prompt than the original input.}
	\label{fig:bert_attention}
\end{figure}

Prompt-based approaches have shown success in low-data regimes \citep{Petroni2019LanguageMA,Schick2021ItsNJ,Jiang2020HowCW,Gao2021MakingPL,Lester2021ThePO}. Prompt-based fine-tuning is beneficial in few-shot learning, because it provides extra task information to the model through the prompt text \citep{Schick2021ItsNJ}. However, when we explore this technique in practice, two issues have arisen. First, the trained model can overfit on words or phrases within the prompt and on the position of the [MASK] token in the prompt \citep{Zhong2021AdaptingLM}. For example, in movie review sentiment analysis, when we append the prompt, ``Does the user like the movie? [MASK]'', to a negative review, ``This is a bad movie.'', the trained model is inclined to predict the positive class, because the word ``like'' frequently appears in positive reviews and the masked language model has greater attention on the words/phrases that are closer to the mask token as shown in Figure~\ref{fig:bert_attention}. We call this issue prompt-related overfitting in this work.

We tackle prompt-related overfitting by introducing a dynamic prompt approach. In this approach, we create a prompt pool consisting of multiple prompt templates. To construct this pool, we generate a set of prompt candidates and filter by a similarity score we propose, called the pairwise prompt dissimilarity score (detailed in Section~\ref{sec:method}). We then introduce the dynamic component of the algorithm by randomly selecting a prompt template from the pool and applying to the input for each training step. For example, in the movie review sentiment analysis task, the trained model will randomly see either ``does the user like the movie? [MASK]'' or ``does the user dislike the movie? [MASK]'' appended to the original input. This prevents the model to overfit on spurious correlations between words in the prompt and the class label.

\begin{figure*}[ht]
	\centering
	\includegraphics[width=.75\linewidth]{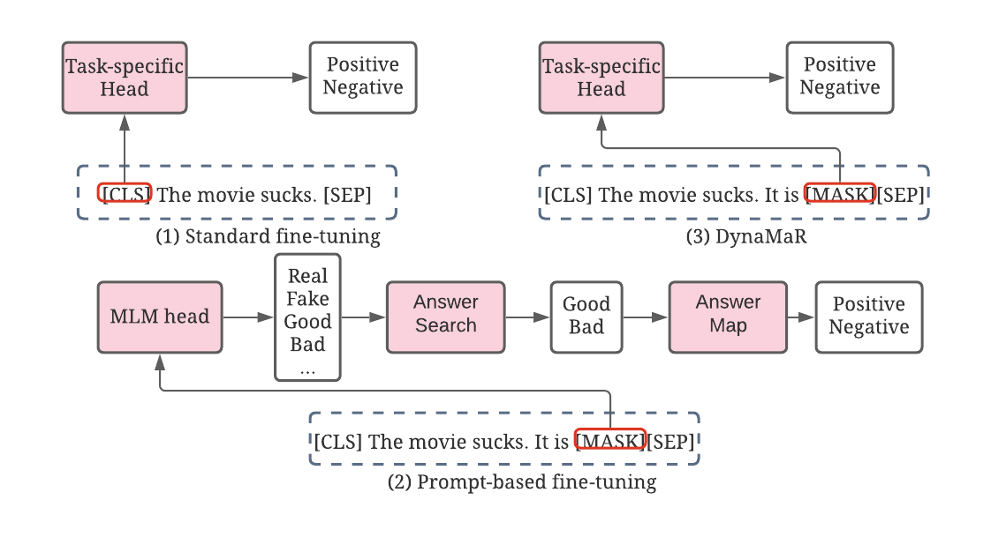}
	\caption{Fine-tuning approach demonstration.}
	\label{fig:finetune_approach}
\end{figure*}

In addition, as previously mentioned, the standard prompt-based fine-tuning setup can be inefficient. It requires significant input and answer engineering to reformulate the downstream tasks as MLM problems \citep{Liu2021PretrainPA}. This process is time-consuming especially for tasks with large numbers of classes. Besides, another disadvantage of the standard setup is that it cannot be directly applied to regression problems, as they cannot be easily converted to MLM problems. To simplify this process, we fine-tune the model by feeding the mask token representation into a task-specific classifier/predictor head instead of the pre-trained MLM head to avoid the answer engineering process, as shown in Figure~\ref{fig:finetune_approach}. We refer to our prompt-based approach with these two improvements as Dynamic Prompt with Mask Token Representation (DynaMaR). We apply DynaMaR to both few-shot and data-rich settings and, for the first time, show improvement gains across four tasks not only in few-shot settings but also in data-rich settings.

Our contributions include: (1) proposing DynaMaR, which can be applied without reformulating downstream tasks into language problems and is robust to prompt-related overfitting, (2) showing DynaMaR can achieve improvements in both few-shot and data-rich settings, (3) proposing a prompt dissimilarity score to evaluate the degree of dissimilarity between two prompt templates and to help construct a diverse dynamic prompt pool, (4) demonstrating that a larger dynamic prompt pool achieves better performance on downstream tasks.

\section{Related Work}

Our work can be divided into three components: language model fine-tuning, prompt generation, and the design of the prompt template.

{\bf Language Model Fine-tuning} is the main focus of our work. Recently, a large amount of research has focused on improved language model finetuning methods \citep{Howard2018UniversalLM,Dodge2020FineTuningPL,Lee2020MixoutER,Zhang2021RevisitingFB}. These works mainly focus on optimization and regularization techniques to stabilize fine-tuning. In contrast to these works, \citet{Gao2021MakingPL} describe the concept of prompt-based fine-tuning for language models.  We adapt and simplify the core ideas from this work to create a simple yet efficient prompt-based fine-tuning approach.

{\bf Prompt Generation} is a key process in prompt-based fine-tuning. The choice of prompt significantly influences performance. The most natural way to generate prompts is through manual design. \citet{Petroni2019LanguageMA} employ manually generated prompts with ELMo \citep{Peters2018DeepCW} and BERT \citep{Devlin2019BERT} models. They evaluate on the LAMA (LAnguage Model Analysis) benchmark \citep{Bordes2013TranslatingEF,Nickel2016ARO} without fine-tuning and conclude that the model is able to recall knowledge learned from the pre-training tasks. While manually crafting prompts is intuitive, creating effective prompts through manual effort requires time, experience, and expertise. To address this issue, a number of automatic prompt searching methods have been proposed. For example, \citet{Jiang2020HowCW} propose a data mining-based method that searches for a prompt based on the shortest path between the original inputs and answers. They also propose paraphrasing-based methods that take a seed prompt and paraphrase it into several semantically similar expressions. \citet{Gao2021MakingPL} treat prompt generation as a text generation task and utilize T5, a sequence-to-sequence pretrained model, in the template search process. They generate templates by specifying the position to insert a prompt template and then inputting samples into T5 to decode the templates. These automatic approaches achieve comparable performance to manually designed prompts. Besides, \citet{LoganIV2021CuttingDO} propose the null prompt method. Instead of generating prompts, they concatenate a [MASK] token with original inputs and it performs competitively to manually designed prompts. In our experiments, we utilize the prompt generation methods to create candidates for the dynamic prompt pool, while also including the null prompt approach as one of the baselines.

{\bf Prompt Template Design Factors} are the factors that we take into consideration to create a metric that informs how prompts are selected for the dynamic prompt pool. Numerous previous works analyze prompt template design factors and the impact of prompt design on performance. \citet{Liu2021PretrainPA} summarize the factors that influence the application of prompt-related technologies in language models. \citet{LoganIV2021CuttingDO} note that the order in which the original input and the [MASK] token are concatenated is an important consideration. \citet{Zhong2021AdaptingLM} propose to unify the prompts into a question-answering format. These previous works indicate that prompt construction impacts performance. To this end, we hypothesize that diversity in the set of prompt templates is an important factor in the performance of the model and propose a prompt dissimilarity score for measuring diversity.

\section{Our Method: DynaMaR }\label{sec:method}

In this section, we describe details of our approach, DynaMaR. Before explaining the training process, we define two concepts: the dynamic prompt pool and the inference prompt.

{\bf Dynamic Prompt Pool} is a pool of prompt templates from which a prompt template will be randomly selected and applied to the input during training.

{\bf Inference Prompt} is the prompt template used during inference. It is selected from the set of templates in the dynamic prompt pool. In general, it is the prompt template among those in the dynamic prompt pool that can achieve the highest performance in a fixed prompt setting.

We generate the candidates for the dynamic prompt pool and inference prompt through manual generation and paraphrasing-based methods proposed by \citet{Jiang2020HowCW}. However, we do not include all candidates in the dynamic prompt pool. We want to ensure the prompts within a pool are sufficiently diverse so that the model will not overfit on any of them. Therefore, we introduce a prompt dissimilarity score to measure the level of dissimilarity between these candidates. We consider three factors in developing this metric: (1) prompt position, or whether to append or prepend the prompt to the input or even insert into the middle of pairwise inputs, (2) prompt wording or the prompt word selection, and (3) prompt format, or whether to create prompts in statement format or in the question-answering format proposed by \citet{Zhong2021AdaptingLM}. To define the prompt dissimilarity score, we first introduce the normalized Hamming distance and the normalized Levenshtein distance.

{\bf Normalized Hamming Distance} is equal to the number of different bits between two binary representations divided by the length of the binary representations \citep{Norouzi2012HammingDM}. Let $HD(b_i,b_j)$ be the Hamming distance between binary representations $b_i$ and $b_j$ with equal length $K$. The equation of normalized Hamming distance $NHD(b_i,b_j)$ then follows: 
\begin{align}
	HD(b_i,b_j) & = \sum_{k=1}^{K} | b_{ik} - b_{jk} |, \\
	NHD(b_i,b_j) & = HD(b_i,b_j) / K.
\end{align}

{\bf Normalized Levenshtein Distance} is equal to the minimum number of operations (substitution, insertion and deletion) required to transform a given string into another string divided by the length of the longer string and is calculated in a recursive fashion \citep{Yujian2007ANL}. Let $LD(s_i,s_j)$ be the Levenshtein distance between string $s_i$ and $s_j$. Let $|s_i |$ and $|s_j |$ be the length of prompt string $s_i$ and $s_j$, respectively. Let $t(x)$ be a function that keeps a string of all but the first character of $x$. The equation of the normalized Levenshtein distance $NLD(s_i,s_j)$ follows:

{\small \begin{align}
	& LD(s_i,s_j) = \begin{cases}
		|s_i|, & \text{if } |s_i| = 0; \\
		|s_j|, & \text{if } |s_j| = 0; \\
		LD(t(s_i), t(s_j)), & \text{if } |s_i| = |s_j|; \\
		1 + \\
		\min \left( \substack{ LD(t(s_i),s_j), \\ LD(s_i,t(s_j)), \\ LD(t(s_i),t(s_j)) } \right), & \text{otherwise.}
	\end{cases} \\
	& NLD(s_i,s_j) = \begin{cases}
		\frac{LD(s_i,s_j)}{|s_i|}, & \text{if } |s_j| \le |s_i|, \\
		\frac{LD(s_i,s_j)}{|s_j|}, & \text{if } |s_i| < |s_j|.
	\end{cases}
\end{align}}

Suppose we generate $N$ prompt templates. Let $p_i$ and $p_j$ be two prompt templates with $s_i$, $s_j$ as prompt strings, respectively, where $i\ne j$ and $i,j\in \{1,2,\dots,N\}$. We treat the prompt position and format information as categorical variables and convert them into binary representations, $b_i$, $b_j$. Let $PDS (p_i,p_j)$ denote the prompt dissimilarity score between prompt templates $p_i$ and $p_j$. The prompt dissimilarity score equation can be found below: \begin{equation}
	PDS(p_i,p_j) = NHD(b_i,b_j) + NLD(s_i,s_j).
\end{equation}

In our experiment, we use 0.5 as the pairwise prompt dissimilarity score threshold. We add the prompt templates that have prompt dissimilarity score larger than the threshold to others to a dynamic prompt pool. During the training process, we randomly pick one prompt template from the pool for each training step and apply it to the original input. We treat the mask token representation from the modified input as the sentence embedding and train the model by directly feeding it into a task-specific predictor head.

\section{Experiment}

\subsection{Data}\label{subsec:data}

In this experiment, we use four e-commerce proprietary datasets: (1) Variation Elimination (VE), (2) Music Match (MM), (3) Music Genre (MG), and (4) Price Prediction (PP). VE is a binary classification problem with pairwise-document inputs where the label identifies whether two items are the variations of the same product or not. For example, similar shirts (from the same producer and brand) in different sizes or colors are considered to be variations. MM is a binary classification problem with pairwise-document inputs that identifies whether two music tracks from different sources are the same or not. MG is a 303-way classification problem with single-document inputs that classifies music tracks to genres. PP is a regression problem with single-document inputs that aims to estimate the sales price based on the product information. It should be noted that the percentage of inputs with number of tokens larger than 512 in VE, MM, MG, PP are 90\%, 75\%, 82\%, 1\%, respectively.

For each task, we split the dataset into three parts: (1) train, (2) validation, and (3) test. We use the full training dataset for the data-rich settings. We also sample multiple few-shot training datasets for few-shot learning settings. In few-shot learning, each classification dataset contains roughly 20 samples for each class. For the regression task (i.e., PP), we randomly sample 1\% of the full training dataset as a few-shot training dataset.

\subsection{Model and Tokenizer Setup}

For training the tokenizer, we collect an English product catalog dataset with text features including title, description, and detail bullet points. We train a 32K BPE vocabulary on this dataset using the SentencePiece library \citep{Kudo2018SentencePieceAS}.

We create a 500M parameter transformer encoder-only model, with 38 hidden layers, 1024 embedding size, 16 attention heads, and maximum sequence length of 512. We train the model using the LANS optimizer \citep{Zheng2020AcceleratedLB} with a batch size of 8192 and a learning rate of $10^{-4}$ on the product catalog dataset.

\subsection{Prompt Generation and Selection}

To create the dynamic prompt pool for our tasks, we first generate 20 prompt templates for each task and select 5 out of them using the prompt dissimilarity score. Specifically, for each task, we first manually design 10 prompt templates. By treating prompt template generation as paraphrase generation task \citep{Jiang2020HowCW}, we use these 10 prompt templates as seeds to generate another 10 templates per task by leveraging the public T5 paraphrase generation model from Hugging Face\footnote{\url{https://huggingface.co/Vamsi/T5_Paraphrase_Paws}}. Afterwards, we use the prompt dissimilarity score to select 5 prompt templates out of the 20 based on the method discussed at the end of Section~\ref{sec:method}. The selected prompt templates are used as each task's dynamic prompt pool. For inference, we evaluate each template in the dynamic prompt pool through the evaluation process discussed in Section~\ref{subsec:model_training}, and select the prompt template that produces the best performance on each task. Table~\ref{tab:pool_size_5} shows the dynamic prompt templates as well as the inference prompt selected for each task.

\subsection{Fine-tuning (Ft) Methods}

We compare DynaMaR with the following approaches:
\begin{itemize}
	\item {\bf Promptless Fine-tuning - CLS (PFt-CLS)} is our baseline approach where we fine-tune the model by feeding the [CLS] token representation into a predictor head.
	\item {\bf Promptless Fine-tuning - Average Pooling (PFt-Avg)} fine-tunes the model by using the average of sequence output for prediction.
	\item {\bf Null Prompt - Prefix (NP-Prefix)} prepends the [MASK] token to the original inputs and fine-tunes the model by feeding the [MASK] token representation into a predictor head. This approach avoids the issue where the model overfits the prompt template since it does not require a template.
	\item {\bf Null Prompt - Suffix (NP-Suffix)} is the same as the above approach except that the [MASK] token is appended to the inputs instead of being prepended.
	\item {\bf Fixed Prompt with Mask Token Representation (FiTeR)} utilizes a static prompt template in both the training and inference stages and fine-tunes the model by feeding the [MASK] token representation into a predictor head. 
\end{itemize}

Note that we use a task-specific predictor head in combination with all above approaches including the prompt-based fine-tuning methods, which typically use the pre-trained MLM head for prediction. The reason is that we have a regression task as one of our evaluation datasets, and as already discussed in Section~\ref{sec:introduction}, it is not straight forward to convert regression tasks into MLM tasks.

\subsection{Model Training and Evaluation Setup}\label{subsec:model_training}

As mentioned in Section~\ref{sec:introduction}, we measure the performance in both few-shot and data-rich settings. For both VE and MM, we use Area Under the Precision-Recall Curve (PRAUC) as the evaluation metric. For MG, we use classification accuracy as the evaluation metric. For PP, we use Root Mean Square Error (RMSE) as the evaluation metric. We validate the performance every 2 training steps in the few-shot  settings and every 100 steps in the data-rich settings. We use early stopping with a patience of 3 validation steps to select the best model for each task. We then evaluate the best models on the test datasets. For few-shot learning, we report the average performance across multiple few-shot datasets per task to reduce the variation in performance. In Table~\ref{tab:results_few_shot} and Table~\ref{tab:results_data_plenty}, we calculate and report the improvement percentage, which is the ratio of positive change as compared to PFt performance.

\subsection{Results}\label{subsec:results}

\bgroup
\def\arraystretch{1.2}
\begin{table}
	\small
	\centering
	\begin{adjustbox}{max width=\linewidth,center}
		\begin{tabular}{lllll | l}
			\hline
			{\bf Ft Method} & {\bf VE} & {\bf MM} & {\bf MG} & {\bf PP} & {\bf Avg} \\
			\hline
			PFt-CLS & 0 & 0 & {\bf 0} & 0 & 0 \\
			PFt-Avg & -1.5\% & +7.2\% & -3.7\% & -8.8\% & -1.7\% \\
			NP-Prefix & -1.0\% & +4.1\% & -2.0\% & +2.6\% & +0.9\% \\
			NP-Suffix & -2.6\% & +0.2\% & -1.6\% & +6.7\% & +0.7\% \\
			FiTeR & -0.7\% & +13.9\% & -1.1\% & +7.3\% & +4.9\% \\
			DPMR & {\bf +0.8\%} & {\bf +15.8\%} & -0.5\% & {\bf +23.8\%} & {\bf +10.0\%} \\
			\hline
		\end{tabular}
	\end{adjustbox}
	\caption{Few-shot Learning Performance Comparison.}
	\label{tab:results_few_shot}
\end{table}
\egroup

\bgroup
\def\arraystretch{1.2}
\begin{table}
	\small
	\centering
	\begin{adjustbox}{max width=\linewidth,center}
		\begin{tabular}{lllll | l}
			\hline
			{\bf Ft Method} & {\bf VE} & {\bf MM} & {\bf MG} & {\bf PP} & {\bf Avg} \\
			\hline
			PFt-CLS & {\bf 0} & 0 & {\bf 0} & 0 & 0 \\
			PFt-Avg & -0.1\% & +1.2\% & -1.0\% & -11.0\% & -2.7\% \\
			NP-Prefix & -0.1\% & +1.0\% & -0.4\% & 0 & +0.1\% \\
			NP-Suffix & -0.3\% & +1.7\% & -0.7\% & +2.2\% & +0.7\% \\
			FiTeR & {\bf 0} & +1.5\% & -0.2\% & +3.3\% & +1.2\% \\
			DPMR & {\bf 0} & {\bf +2.9\%} & -0.3\% & {\bf +12.1\%} & {\bf +3.7\%} \\
			\hline
		\end{tabular}
	\end{adjustbox}
	\caption{Data-rich Performance Comparison.}
	\label{tab:results_data_plenty}
\end{table}
\egroup

Table~\ref{tab:results_few_shot} and \ref{tab:results_data_plenty} show the performance results for both few-shot and data-rich settings. In both settings, PFt-Avg shows degradation in average of performance compared to PFt-CLS. This shows that average pooling generates worse sentence representations than does taking the [CLS] token representation.

In contrast, both null prompt approaches show improvement in average performance compared to PFt-CLS in both few-shot and data-rich settings. The improvement could be a result of aligning the format of the downstream tasks and that of the pre-training task. By changing the input format to be similar to that of the MLM task, we reduce the amount of data that are required to coach the model to learn the new task.

Also, there is a difference in the performance of NP-suffix and NP-prefix. This is likely due to the positional differences of the [MASK] token in the two methods. For example, suppose we want to perform sentiment analysis on a sentence like ``I love the movie''. Prepending or appending the [MASK] token would result in different distances between [MASK] and the word ``love'', which holds the key information for classification. Such positional differences could lead to different performance even though the two methods are very similar in spirit.

Another observation is that FiTer shows higher improvement in average of performance compared to null prompt approaches. Recall that FiTer introduces task information through the prompt templates, while the null prompt approaches do not, which supposedly addresses the issue where the model overfits the prompt templates. Hence, the results show that the benefits of adding the extra task information outweigh the possible performance loss caused by the prompt-related overfitting issue.

Finally, DynaMar outperforms FiTer on all tasks in both setting, with the only exception being MG in the data-rich setting. This indicates that increasing the diversity of prompt templates used during training will improve model generalization. We also observe that DynaMar does not show significant improvement over PFt-CLS on both MG and VE. This is because both tasks contain a large number of documents with length longer than 512, as mentioned in Section~\ref{subsec:data}. As a result of this, we need to truncate more of the original inputs for these tasks in order to insert prompts, which can lead to information loss. Thus, DynaMar is less efficient in problems with long documents.

\subsection{Analysis}

{\bf Larger dynamic prompt pool, better performance.} The size of the dynamic prompt pool influences the performance. We compare the average improvement percentage across four tasks with the size of dynamic prompt pool = 1, 3, 5 (prompt information can be found in Appendix~\ref{sec:dynamic_prompt_pool}). From Figure~\ref{fig:pool_size_exp}, we can see that performance improves as the dynamic prompt pool is made larger.

\begin{figure}[h]
	\centering
	\includegraphics[width=.85\linewidth]{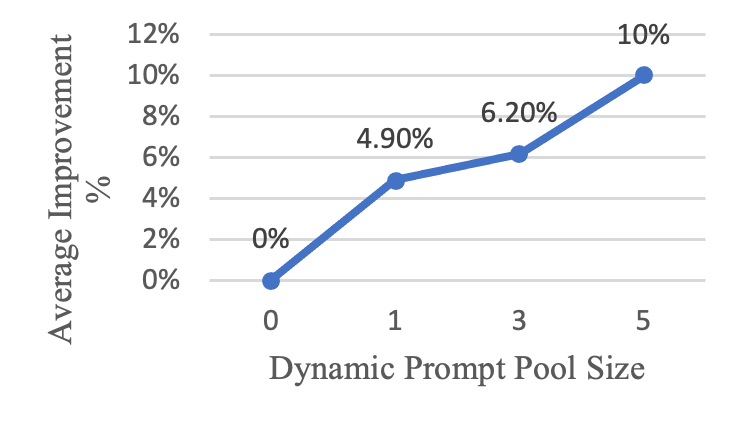}
	\caption{Pool Size vs Improvement Percentage.}
	\label{fig:pool_size_exp}
\end{figure}

\subsection{Limitations and Future Directions}

As mentioned in Section~\ref{subsec:results}, our method does not show substantial improvement on tasks involving long documents. Besides, the threshold of prompt disimilarity score can be treated as a parameter. This work lack of a study on the effect of this threshold. In addition, we focus on e-commerce related English classification/regression tasks in this work, the performance of our method in other nature language processing use cases remains unexplored. As a next step, we will conduct additional studies on these three topics.

\section{Conclusion}

In this work, we discuss methods for generating prompts and propose a way to select prompt templates to include in the dynamic prompt pool. Also, we show that using the mask representation of a prompt either equals or improves upon the performance of standard fine-tuning on four e-commerce applications in both few-shot and data-rich settings. In addition, we find DynaMaR outperforms the fixed prompt approach in both settings. Furthermore, we show that a larger dynamic prompt pool leads to improved model performance when employing DynaMaR.

\bibliography{anthology,custom}
\bibliographystyle{acl_natbib}

\appendix

\section{Dynamic Prompt Pool with Different Sizes}\label{sec:dynamic_prompt_pool}

We need to define two prompt-related parameters while using DynaMaR: the dynamic prompt pool and the inference prompt. The list of prompts in the pool and the inference prompt selected for dynamic prompt pool sizes of 1, 3, and 5 can be found in Table~\ref{tab:pool_size_1}, Table~\ref{tab:pool_size_3}, and Table~\ref{tab:pool_size_5}, respectively.

\bgroup
\def\arraystretch{1.4}
\begin{table*}[t]
	\small
	\centering
	\begin{tabular}{lll}
		\hline
		{\bf Task} & {\bf Inference Prompt} & {\bf Dynamic Prompt Pool} \\
		\hline
		VE & $f(x_1, x_2)$ = $x_1$ and $x_2$ are [MASK] product & $f(x_1, x_2)$ = $x_1$ and $x_2$ are [MASK] product \\
		\hline
		MM & $f(x_1, x_2)$ = $x_1$ and $x_2$ are [MASK] music & $f(x_1, x_2)$ = $x_1$ and $x_2$ are [MASK] music \\
		\hline
		MG & $f(x)$ = Genre: [MASK] $x$ & $f(x)$ = Genre: [MASK] $x$ \\
		\hline
		PP & $f(x)$ = $x$ The price is [MASK] & $f(x)$ = $x$ The price is [MASK] \\
		\hline
	\end{tabular}
	\caption{Dynamic Prompt Pool Size = 1.}
	\label{tab:pool_size_1}
\end{table*}
\egroup

\bgroup
\def\arraystretch{1.4}
\begin{table*}[t]
	\small
	\centering
	\begin{tabular}{lll}
		\hline
		{\bf Task} & {\bf Inference Prompt} & {\bf Dynamic Prompt Pool} \\
		\hline
		VE & $f(x_1, x_2)$ = $x_1$ and $x_2$ are [MASK] product & \makecell{
			(1) $f(x_1, x_2)$ = $x_1$ $x_2$. Are they the same product? [MASK] \\
			(2) $f(x_1, x_2)$ = $x_1$ and $x_2$ are [MASK] product \\
			(3) $f(x_1, x_2)$ = $x_1$ $x_2$. They are [MASK]
		} \\
		\hline
		MM & $f(x_1, x_2)$ = $x_1$ and $x_2$ are [MASK] music & \makecell{
			(1) $f(x_1, x_2)$ = $x_1$ $x_2$. Are they the same song? [MASK] \\
			(2) $f(x_1, x_2)$ = $x_1$ and $x_2$ are [MASK] music \\
			(3) $f(x_1, x_2)$ = $x_1$ is as [MASK] as $x_2$
		} \\
		\hline
		MG & $f(x)$ = Genre: [MASK] $x$ & \makecell{
			(1) $f(x)$ = Genre: [MASK] $x$ \\
			(2) $f(x)$ = Music Genre: [MASK] $x$ \\
			(3) $f(x)$ = $x$ what is genre of the music? [MASK]
		} \\
		\hline
		PP & $f(x)$ = $x$ The price is [MASK] & \makecell{
			(1) $f(x)$ = Price: [MASK] $x$ \\
			(2) $f(x)$ = $x$ it cost [MASK] dollars \\
			(3) $f(x)$ = $x$ what is price of the product? [MASK]
		} \\
		\hline
	\end{tabular}
	\caption{Dynamic Prompt Pool Size = 3.}
	\label{tab:pool_size_3}
\end{table*}
\egroup

\bgroup
\def\arraystretch{1.4}
\begin{table*}[t]
	\small
	\centering
	\begin{tabular}{lll}
		\hline
		{\bf Task} & {\bf Inference Prompt} & {\bf Dynamic Prompt Pool} \\
		\hline
		VE & $f(x_1, x_2)$ = $x_1$ and $x_2$ are [MASK] product & \makecell{
			(1) $f(x_1, x_2)$ = $x_1$ $x_2$. Are they the same product? [MASK] \\
			(2) $f(x_1, x_2)$ = $x_1$ and $x_2$ are [MASK] product \\
			(3) $f(x_1, x_2)$ = $x_1$ $x_2$. They are [MASK] \\
			(4) $f(x_1, x_2)$ = Are $x_1$ and $x_2$ the same product? [MASK] \\
			(5) $f(x_1, x_2)$ = $x_1$ is as [MASK] as $x_2$
		} \\
		\hline
		MM & $f(x_1, x_2)$ = $x_1$ and $x_2$ are [MASK] music & \makecell{
			(1) $f(x_1, x_2)$ = $x_1$ $x_2$. Are they the same song? [MASK] \\
			(2) $f(x_1, x_2)$ = $x_1$ and $x_2$ are [MASK] music \\
			(3) $f(x_1, x_2)$ = $x_1$ $x_2$. They are [MASK] music \\
			(4) $f(x_1, x_2)$ = Are $x_1$ and $x_2$ the same music? [MASK] \\
			(5) $f(x_1, x_2)$ = $x_1$ is as [MASK] as $x_2$
		} \\
		\hline
		MG & $f(x)$ = Genre: [MASK] $x$ & \makecell{
			(1) $f(x)$ = Genre: [MASK] $x$ \\
			(2) $f(x)$ = Music Genre: [MASK] $x$ \\
			(3) $f(x)$ = $x$ This is a [MASK] music \\
			(4) $f(x)$ = Type: [MASK] $x$ \\
			(5) $f(x)$ = $x$ what is genre of the music? [MASK]
		} \\
		\hline
		PP & $f(x)$ = $x$ The price is [MASK] & \makecell{
			(1) $f(x)$ = Price: [MASK] $x$ \\
			(2) $f(x)$ = $x$ Price: [MASK] \\
			(3) $f(x)$ = $x$ it cost [MASK] dollars \\
			(4) $f(x)$ = $x$ The price is [MASK] \\
			(5) $f(x)$ = $x$ what is price of the product? [MASK]
		} \\
		\hline
	\end{tabular}
	\caption{Dynamic Prompt Pool Size = 5.}
	\label{tab:pool_size_5}
\end{table*}
\egroup

\end{document}